\documentclass[10pt,journal,compsoc]{IEEEtran}

\ifCLASSOPTIONcompsoc
  \usepackage[nocompress]{cite}
\else
  \usepackage{cite}
\fi
\ifCLASSINFOpdf
  \usepackage[pdftex]{graphicx}
\else
\fi
\usepackage{amsfonts}
\usepackage{amsmath}
\usepackage{amssymb}
\usepackage{multirow}
\usepackage{scalerel,stackengine}
\usepackage{subfigure}
\usepackage{array}
\begin{document}
%
\title{EvAn: Neuromorphic Event-based Anomaly Detection}
%
%
%
%

\author{Lakshmi Annamalai,
        Anirban Chakraborty, {Member,~IEEE},
        and~Chetan Singh Thakur, {Senior Member,~IEEE}.
\IEEEcompsocitemizethanks{\IEEEcompsocthanksitem Lakshmi Annamalai is with Defence Reserach and Development Organization, Bangalore, India. This work was done at NeuRonICS lab, Department of Electronic Systems Engineering, Indian Institute of Science, Bangalore, India, where Lakshmi is currently pursuing her doctoral research\protect\\
E-mail: lakshmia@iisc.ac.in
\IEEEcompsocthanksitem Anirban Chakraborty is with Department of Computational and Data Sciences at Indian Institute of Science, Bangalore, India.\protect\\
E-mail: anirban@iisc.ac.in
\IEEEcompsocthanksitem Chetan Singh Thakur is with Department of Electronic Systems Engineering at Indian Institute of Science, Bangalore, India.\protect\\
E-mail: csthakur@iisc.ac.in}
}

\IEEEtitleabstractindextext{%
\begin{abstract}
Event-based cameras are bio-inspired novel sensors that asynchronously record changes in illumination in the form of events, thus resulting in significant advantages over conventional cameras in terms of low power utilization, high dynamic range, and no motion blur. Moreover, such cameras, by design, encode only the relative motion between the scene and the sensor (and not the static background) to yield a very sparse data structure, which can be utilized for various motion analytics tasks. In this paper, for the first time in event data analytics community, we leverage these advantages of an event camera towards a critical vision application - video anomaly detection. We propose to model the motion dynamics in the event domain with dual discriminator conditional Generative adversarial Network (cGAN) built on state-of-the-art architectures. To adapt event data for using as input to cGAN, we also put forward a deep learning solution to learn a novel representation of event data, which retains the sparsity of the data as well as encode the temporal information readily available from these sensors. Since there is no existing dataset for anomaly detection in event domain, we also provide an anomaly detection event dataset with an exhaustive set of anomalies. Careful analysis reveals that the proposed method results in huge reduction in computational complexity as compared to previous state-of-the-art conventional anomaly detection networks. 
\end{abstract}

\begin{IEEEkeywords}
Neuromorphic Camera, Event data, Anomaly Detection, Generative Adversarial Network.
\end{IEEEkeywords}}

\maketitle

\IEEEdisplaynontitleabstractindextext

%
\IEEEpeerreviewmaketitle

\ifCLASSOPTIONcompsoc
\IEEEraisesectionheading{\section{Introduction}\label{sec:introduction}}
\else
\section{Introduction}
\label{sec:introduction}
\fi

\begin{figure}
\centering
\includegraphics[width=\linewidth,height=1.5 in]{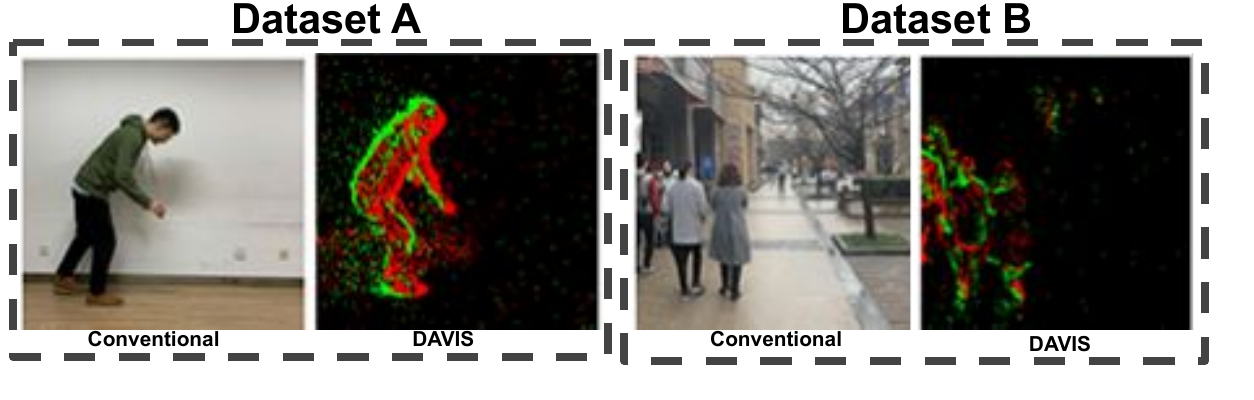}  
\caption{Display of images captured by conventional camera and DAVIS as provided by \cite{MiaoShuandChenGuang}. These are part of DAVIS dataset (Dataset A and B, details of which are given in Section. 4.1) that has been ventured in this paper for anomaly detection.}
\label{fig:disk}
\end{figure}

This paper focusses on anomaly detection using bio-inspired event-based cameras that register pixel-wise changes in brightness asynchronously in an efficient manner, which is radically different from how a conventional camera works. This results in a stream of events (Fig. \ref{fig:disk}) $e_k$, where $e_k=\{x_k,y_k,t_k,p_k\}$, $x_k$ and $y_k$ being the x and y coordinates of the pixel where an intensity change of pre-defined threshold has happened, $t_k$ and $p_k$ are the time (accurate to microseconds) and polarity $\in\{+1,-1\}$ of the change. The asynchronous principle of operation endows event cameras \cite{TDelbruckBLinaresBarrancoECulurciello} \cite{TDelbruckandCMead} \cite{CPoschandDMatolinandRWohlgenannt} \cite{TSerranoGotarredonaandBLinaresBarranco} to capture high-speed motions (with temporal resolution in the order of ${\mu}s$), high dynamic range ($140db$) and sparse data. These low latency sensors have paved way to develop agile robotic applications \cite{LakshmiAandAnirbanChakraborty}, which was not feasible with conventional cameras. Only limited achievements have been accomplished in designing robust and accurate visual analytics algorithms for the event data, mainly because of unavailability of event cameras for commercial purposes and resultantly, dearth of large scale event datasets.

Video anomaly detection \cite{JCSanMiguelandJMMartinez} \cite{SWJooandRChellappa} is a pervasive application of computer vision with its widespread applications as diverse as surveillance, intrusion detection etc. Anomaly detection can be posed as a foreground motion analytics task. This makes event camera an ideal candidate for video anomaly detection task as it comes with the ability to encode motion information at sensor level and provides automatic background subtraction for stationary camera surveillance \ref{fig:conv_davis}. We argue that a vast majority of state-of-the-art frame-based anomaly detection networks rely on optical flow estimation or $3D$ convolution to explicitly model motion constraint to detect anomalies. It is pretty important to note that event cameras exhibits embedded motion information at the sensor level, which allows us to circumvent optical flow or $3D$ convolution. Hence, direct application of existing conventional anomaly detection networks to event data is debatable.

In this paper, we introduce a solution towards solving event-based anomaly detection with a dual discriminator conditional Generative adversarial Network (cGAN), which has not been explored in prior works of event vision. Dual discriminator allows detection of spatial event anomalies, which would have not been feasible otherwise. However, data modality of event data does not directly fit this network due to its inherent difference from conventional cameras. State-of-the-art deep learning compatible event representations belong to feature engineering which mandates domain expertise. However, the advantages of deep learning to learn nested concepts from data has not been explored in event data representation. To alleviate this gap, we introduce a less computationally complex shallow encoder-decoder architecture (referred here as DL (Deep Learning) memory surface generation network) that learns a better latent-space sparse event representation while efficiently modeling motion analytics. To validate the efficacy of our algorithm, we introduce for the first time, a novel anomaly detection event dataset recorded from a type of event camera known as Dynamic Active Pixel Vision Sensor (DAVIS) \cite{MoeysanDiederikPaulandFederico} \cite{PatrickLichtsteinerandChristophPoschandTobiDelbruck}. 

\textbf{Contributions: }In the context of the previous discussion, our contributions in this paper can be summarized as,
\begin{enumerate}
\item{Event Data Representation: Shallow encoder-decoder module to generate an appropriate sparse representation of event data that inherently learns the temporal information from the data as opposed to state-of-the-art handcrafted features}

\item{Anomaly detection network: Dual discriminator cGAN that combines the conditional and unconditional setting of the discriminator (inspired from \cite{IsolaPandZhuJY} and \cite{NguyenTandLeTandVuH}}) to model spatial and temporal anomalies in event domain.

\end{enumerate}

\section{Related Work: }
In this section, we review previous work in the areas of event data representation and conventional frame-based anomaly detection.

\subsection{Event data representation: }The models that can cope with event data are biologically inspired neural networks acknowledged as Spiking Neural Networks (SNN) \cite{ARussellandGOrchardandYDong}. SNN has not become increasingly popular due to the lack of scalable training procedures. An alternative methodology followed in literature is an adaptation of event data to make them compatible with conventional networks. 

Earlier works restrained themselves to encoding basic information such as polarity. In \cite{ANguyenTTDoDGCaldwellNGTsagarakis}, list of events are converted into images by recording the occurrence of the recent event in the given time. The drawback of this representation is that it encodes solely the latest event information at each pixel value. In \cite{MaquedaAIandLoquercioAandGallegoG}, a two-channel event image is created with histogram $h^{+}$ and $h^{-}$ of positive and negative events respectively. Storage of events of different polarity in different channels avoids the cancellation of events of opposite polarity at a given location. This choice proves to be better than that of \cite{ANguyenTTDoDGCaldwellNGTsagarakis}. The predominant setback of the above basic strategies is that they discard the treasured time information that is obtained from event cameras.

Non-spatial time encodings contain useful motion information, including which will enhance the accuracy of vision algorithms. However, incorporating this information is a cumbersome task. In \cite{PaulKJParkBaekHwanChoJinManPark}, time-stamp maps are created by using three distinctive techniques, pixel replication, temporal interpolation, and spatio-temporal correlation. These time-stamp maps are merged temporally for further processing, hence tending to lose the details of the time information obtained from the event camera. 

In \cite{XavierLagorceandGarrickOrchard}, the intensity image has been coded with the timestamp of each pixel $\left(x,y\right)$ of recent positive and negative events in the given integration time $T$ and around a spatial location of $R{\times}R$. This image was further used to construct features recognized as time surfaces. Following this, \cite{AZhuandLYuanandKChaney} encode the first two channels as number of positive and negative events that have occurred at each pixel and the last two channels as the time-stamp of the most recent positive and negative events. This representation discards all the other time information except that of the recent event. Moreover, this kind of encoding is very sensitive to noise.

\cite{CYeAMitrokhinCParameshwara} and \cite{AlonsoandInigoandAnaCMurillo} has attempted to improve the time channel information by expertly combining the time information. In \cite{CYeAMitrokhinCParameshwara}, third channel stores average of the timestamp of the events that occurred at pixel $\left(x,y\right)$ in a given temporal window of size $\delta{t}$. \cite{AlonsoandInigoandAnaCMurillo} improved it by allocating four channels that encode standard deviation of the timestamps of positive and negative events (separately) that happened at that specific pixel in the given time interval $\delta{t}$ in addition to their average value.

Recently, \cite{SironiAandBrambillaMandBourdisN} proposed an interesting approach to encode time information by introducing a representation (highly resistant to noise) known as memory surfaces which exponentially weighs the information carried by past events. Following this, \cite{ZhuAZandYuanLandChaneyK} has proposed an event representation by discretizing the time domain. However, this representation might result in higher computational cost when applied on deep network. \cite{CalabreseEandTaverniGandAwaiEasthopeC} has generated frames by accumulating constant number of events, thus claiming to have adaptive frame rate. 

\subsection{Anomaly Detection on Conventional Camera: }As there is no prior work on event data anomaly detection, we briefly describe the frame-based deep learning algorithms for anomaly detection \cite{KiranBandDilipThomas}. Researchers build a statistical model (reconstruction modelling and predictive modelling) to characterize the normal samples and the actions that deviate from the estimated model are identified as anomalies.

Reconstruction modelling \cite{ANg} \cite{MSabokrouMFathyandMHoseini} \cite{RChalapathyandAKMenon} \cite{YSChongandYHTay} \cite{MHasanandJChoiandJNeumann} usually train a deep auto-encoder type neural network to memorize the training videos so that they reconstruct normal events with lesser reconstruction error. It can be presumed that learning capability and generalization of deep networks are too high that they don't conform to the expectation of higher reconstruction error for abnormal events. This led to the new attractive phase of predictive models, which is trained to predict the present frame based on the history of past events. The frames which do not agree with the prediction are declared as anomalies. Researchers have contributed a lot towards predictive modeling with convolutional LSTM \cite{JRMedel} \cite{WLuoWLiuandSGao} \cite{JRMedelandASavakis} and generative modeling. While convolutional LSTM learns the transformation required to predict the frames, generative models such as, Variational Auto-encoder (VAE) \cite{MWDiederikandPKingma} and Generative Adversarial Network (GAN) \cite{GoodfelloandJPougetAbadie}, learn the probability distribution to generate the present from the history, which makes it an ideal candidate for anomaly detection. As this work is based on generative models, we restrict our survey to generative model based conventional anomaly detection.

\cite{TSchleglandPSeebockandSMWaldstein} proposed AnoGAN to detect the manifold of abnormal anatomical variability. AnoGAN is trained with weighted sum of residual loss and discriminator loss. Residual loss, $|x-G\left(z\right)|$, is defined as dissimilarity between original image $x$ and generated image $G\left(z\right)$. Discriminator loss is defined as dissimilarity between intermediate feature representation of original image and that of the reconstructed image, which makes the discriminator as feature extractor, not as hard classifier. However, temporal information has been discarded in modelling anomalies.

\cite{MRavanbakhshandESangineto} proposed an anomaly detection framework that tries to model the anomalies based on motion inconsistency as well. The framework consists of two conditional GAN (cGAN) networks, $\emph{N}^{F\longrightarrow{O}}$ and  $\emph{N}^{O\longrightarrow{F}}$ trained on cross channel tasks of generating $t^{th}$ frame of training video $F_t$ from the optical flow $O_t$ obtained from $F_t$ and $F_{t+1}$ using \cite{TBroxandABruhnandNPapenberg} and vice-versa. During test time, discriminators $D^{F\longrightarrow{O}}$ and $D^{O\longrightarrow{F}}$ identify abnormal areas that correspond to outliers with respect to the distribution learnt by the discriminators during the training phase. 

Similar architecture has been followed in \cite{RavanbakhshMandNabiM}. However the major difference lies in the methodology used to detect possible anomalies. At test time, this work utilizes generators to reconstruct optical image and frame, which results in unstructured blobs for anomalous events due to the inability of the network to reconstruct unseen abnormal events.

\cite{LiuWandLuoWandLianD} proposed a GAN solution to capture appearance and motion based anomalies by leveraging reconstruction loss, gradient loss and optical flow loss in addition to the adversarial loss. The constraint on the motion is modeled as difference between the optical flow of predicted frames and the original frame. Recently, \cite{YanMengjiaandXudongJiangandJunsongYuan} has proposed 3D convolutional GAN to capture temporal information for anomaly detection. $3D$ convolution increases the computational complexity of the network. We argue that frame-based anomaly networks suffer from the weakness of explicitly modeling motion either by optical flow or 3D architectures, which renders it direct application on event data debatable, as the latter exhibits embedded times-tamp information at sensor level.

\section{Proposed Anomaly Detection Method}

\begin{figure*}
\centering
\includegraphics[width=\linewidth,height=2.5 in]{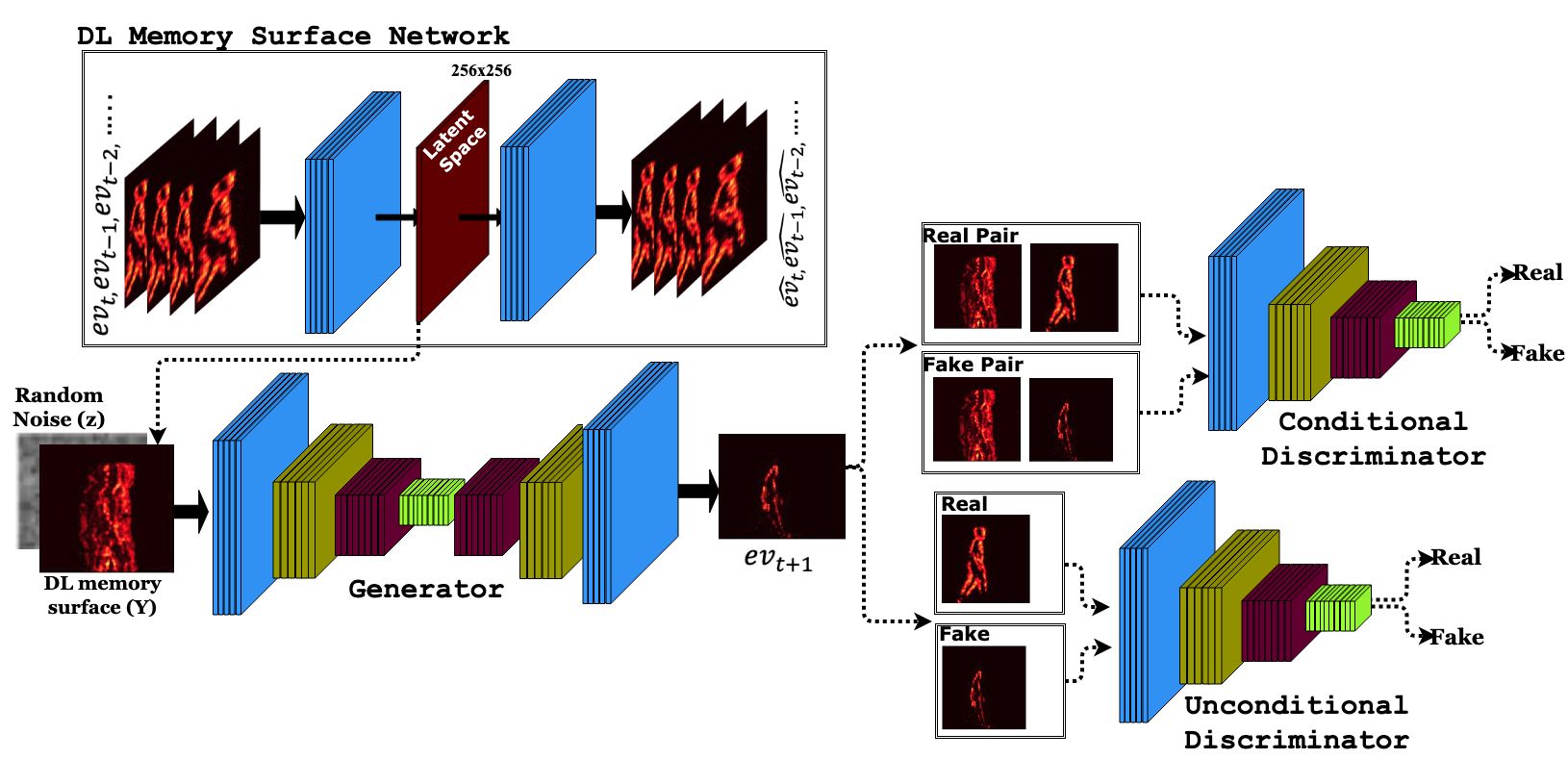}  
\caption{Framework of the proposed DL memory surface network and dual discriminator cGAN. DL memory surface network is trained as stand alone. Dual discriminator cGAN predicts future events conditioned on DL memory surfaces}
\label{fig:framework}
\end{figure*}

The pipeline of our event data prediction framework for anomaly detection is shown in Fig. \ref{fig:framework}. In our methodology, we present anomaly detection as a conditional generative problem that predicts future events conditioned on past events. To predict the future events, we train a dual discriminator conditional GAN with two discriminators (details of which is given in the forthcoming section), one under conditional setting and other under the unconditional setting.  As the generator network of GAN is deep, the solution of predicting future events conditioned on the previous events will confront with computationally heavier model. To make the computation faster, we introduce a DL memory surface generation network (details of which are furnished in upcoming section), that tries to capture the coherence of motion from the given set of events $\{x_i,y_i,p_i,t_i\}_{t_i\in{T}}$ into a single $2D$ structure known as DL memory surface, conditioned on which cGAN learns to predict the future events. To feed the DL memory surface network, we stack the events to produce a discretized volume $Ev = [ev_0,ev_1\ldots{ev_B}]$ (adapted from \cite{ZhuAZandYuanLandChaneyK}), given a time duration $T$ and a set of $B$ discrete time bins $[b_0,b_1\ldots{b_B}]$ each with $\Delta{T}$ duration. 

\subsection{DL Memory Surface Generation Network}

In this work, we propose a novel event representation generated by a shallow, computationally inexpensive encoder-decoder network architecture with a loss function that comprises data term and sparsity term. Before dwelling into the loss function, we introduce the architecture of the encoder-decoder network.

\subsubsection{Network Architecture}

We adapt fully convolutional encoder-decoder architecture \cite{GEHintonandRRSalakhutdinov} with layers of the form convolution (with sigmoid activations) that maps a discretized volume of event data (explained above) to a single image known as DL memory surface $Ms$ with same spatial resolution at the bottleneck layer. The input and output are renderings of the discretized volume data. 

In order to model only the temporal structure in the data without upsetting the spatial distribution, it is adequate if we restrict the convolution operation across the time dimension. Hence, we have constructed the network with $1D$ convolution layers inspired by the paper \cite{LinanMinandQiangChen}, which was the first to introduce $1\times{1}$ convolution. As studies have shown, $1D$ CNN can learn complex tasks  \cite{43022} \cite{FNIandolaandMWMoskewicz}  with shallow architecture unlike its counterpart $2D$ CNN, thereby resulting in "small" network. We have utilized the potential of deep learning to model the motion, thereby capturing full information available from the data that they model in contrast to the state-of-the-art hand-crafted frame generation counterparts.

\subsubsection{Loss Function for Learning Temporal Information}

The DL memory surface generation network tries to learn a function $h_{\theta_{MS}}(Ev)$ such that the target values $[\widehat{ev_0},\widehat{ev_1}\ldots{\widehat{ev_B}}]$ is similar to that of the input $[ev_0,ev_1\ldots{ev_B}]$, while the bottleneck layer models the temporal information encoded by the event camera. The architecture has two parts, encoder and decoder defined by the transformation functions ${\phi}_{\theta_E}:Ev\rightarrow{Ms}$ and $\psi_{\theta_D}:Ms\longrightarrow{Ev}$ respectively with $\theta_D$ and $\theta_E$ being the parameters of decoder and encoder respectively. 

In order to maximize the usefulness of the latent variable encoding, we put forth a data term that tries to model the probability distribution $\{{\mathbb{P}}\left(Ev\mid Ms^*\right)\}$ of getting the event discretized volume, $Ev$, given the ideal DL memory surface, $Ms$, by maximizing the forward $KL$ divergence between the ideal distribution ${\mathbb{P}}\left(Ev\mid Ms^*\right)$ and our estimate ${\mathbb{P}}\left(Ev\mid\widehat{Ms}\right)$ (Eq. \ref{MS_1}). Forward $KL$ divergence will result in best latent variable that covers all the modes of probability distribution of normal videos. 

\begin{eqnarray}
\label{MS_1}
&\mathbb{E}_{Ev\sim{{\mathbb{P}}}\left(Ev\mid Ms^*\right)}\log\left[{{\mathbb{P}}\left(Ev\mid Ms^*\right)}\right]\\ \nonumber
&-\mathbb{E}_{Ev\sim{{\mathbb{P}}}\left(Ev\mid Ms^*\right)}\log\left[{{\mathbb{P}}\left(Ev\mid \widehat{Ms}\right)}\right] \nonumber
\end{eqnarray}

As the first term does not depend on the estimated latent variable, it could be ignored. Hence, the second term of Eq. \ref{MS_1} boils down to maximizing the log likelihood of ${\mathbb{P}}\left(Ev\mid \widehat{Ms}\right)$ when the sample size tends to infinity. The output of the decoder can be modeled as a function of latent variable $\widehat{Ms}$ and noise $\eta\sim\mathbb{•}{N}\left(0,1\right)$ as $\psi_{\theta_D}\left(\widehat{Ms}\right)+\eta$. This makes ${\mathbb{P}}\left(Ev\mid\widehat{Ms}\right)$ a Gaussian distribution with mean $\psi_{\theta_D}\left(\widehat{Ms}\right)$. Thus maximizing the log likelihood turns out to minimizing $-\Vert{{Ev}-\psi_{\theta_D}\left(\widehat{Ms}\right)}\Vert^2$

\subsection{Anomaly Detection Network}

The framework adopted here is a dual discriminator conditional GAN architecture. A striking effect of stationary event camera under surveillance is that it captures moving objects alone, which results in inexpensive modeling of temporal anomalies. However, this has the effect of producing a silhouette of moving objects. Hence, cGAN results in poor characterization of spatial anomalies whose temporal modality overlaps with that of normal data. This could be attributed to the fact that cGAN penalizes the mismatch between input and output by modeling the joint distribution of input and output. In order to incentivize the detection of spatial anomalies in event data, we propose a variant of cGAN which pushes the generator distribution closer to the ground truth distribution while still sustaining the quality of match between input and output

We start this section with a brief review of conditional GAN architecture. For the sake of easy readability, we have used the notation $x$ and $y$ for $ev_{b+1}$ and $\widehat{Ms}$ respectively. Conditional GAN is a two-player game, wherein a discriminator takes two points x and y in data space and emits high probability indicating that they are samples from the data distribution, whereas a generator maps a noise vector z drawn from ${\mathbb{P}}\left(z\right)$ and input sample $y$ drawn from ${\mathbb{P}}_d\left(y\right)$ to a sample $\widehat{x}=G(z,y)$ that closely resembles the data $x$. This is learned by solving the following minimax optimization

\begin{eqnarray}
\label{AD_1}
\min_G\max_D&\mathbb{E}_{y\sim {\mathbb{P}}_d(y)}\mathbb{E}_{x\sim {\mathbb{P}}_d(x\mid y)}\log \left[D\left(x,y\right)\right]\\
&+\mathbb{E}_{y\sim {\mathbb{P}}_d(y)}\mathbb{E}_{\widehat{x}\sim {\mathbb{P}}_g(x\mid y)}\log\left(1-D\left(\widehat{x},y\right)\right] \nonumber
\end{eqnarray}

\textbf{Preposition: }For a fixed G, the optimal discriminator results in 

\begin{equation}
\label{AD_2}
D^*\left(x,y\right)=\frac{{\mathbb{P}}_{dd}(x,y)}{{\mathbb{P}}_{dd}(x,y)+{\mathbb{P}}_{gd}(x,y)}
\end{equation}

With this given $D^*\left(x,y\right)$, the minimization of $G$ turns into minimizing the Jensen-Shannon divergence $D_{JS}=JSD\left[{\mathbb{P}}_{dd}(x,y)||{\mathbb{P}}_{gd}(x,y)\right]$

\textbf{Proof: }Left hand side of Eq.\ref{AD_1} can be expanded as 

\begin{eqnarray}
&\displaystyle  \int_y{{\mathbb{P}}_d(y)\int_x{{\mathbb{P}}_d(x/y)}\log[D(x,y)]}\\
&+\displaystyle  \int_x{{\mathbb{P}}_g(x/y)\int_y{{\mathbb{P}}_d(y)}\log[1-D(x,y)]} \nonumber
\end{eqnarray}

The above equation becomes

\begin{eqnarray}
&\displaystyle  \int_x\int_y{{\mathbb{P}}_{dd}(x,y)}\log[D(x,y)]\\
&+\displaystyle  \int_x\int_y{{\mathbb{P}}_{gd}(x,y)}\log[1-D(x,y)] \nonumber
\end{eqnarray}

The optimal $D^*\left(x,y\right)$ is estimated by differentiating the above equation with respect to $D(x,y)$ and equating to zero,

\begin{equation}
\frac{{\mathbb{P}}_{dd}(x,y)}{D(x,y)}=\frac{{\mathbb{P}}_{gd}(x,y)}{1-D(x,y)}
\end{equation}

On simplification, we get $D^*\left(x,y\right)$ as given in Eq. \ref{AD_2}. Substituting the optimum value of $D$ in the generator equation, it yields

\begin{eqnarray}
\min_G &\displaystyle  \int_x\int_y{{\mathbb{P}}_{dd}(x,y)}\log\left[\frac{{\mathbb{P}}_{dd}(x,y)}{{\mathbb{P}}_{dd}(x,y)+{\mathbb{P}}_{gd}(x,y)}\right]+\\
&\displaystyle  \int_x\int_y{{\mathbb{P}}_{gd}(x,y)}\log\left[\frac{{\mathbb{P}}_{gd}(x,y)}{{\mathbb{P}}_{dd}(x,y)+{\mathbb{P}}_{gd}(x,y)}\right] \nonumber
\end{eqnarray}

Multiplying and dividing the terms inside $\log$ by $2$ and using the fact $\log(AB)=\log(A)+\log(B)$, we get

%
\begin{eqnarray}
\min_G &\lbrace-2\log(2)+KL\left[{\mathbb{P}}_{dd}(x,y)||\frac{{\mathbb{P}}_{dd}(x,y)+{\mathbb{P}}_{gd}(x,y)}{2}\right]\\
&+KL\left[{\mathbb{P}}_{gd}(x,y)]\frac{{\mathbb{P}}_{dd}(x,y)+{\mathbb{P}}_{gd}(x,y)}{2}\right]\rbrace \nonumber
\end{eqnarray}

The second and third term together is nothing but $JSD({\mathbb{P}}_{dd}(x,y)||{\mathbb{P}}_{gd}(x,y))$. Thus, the objective function of $G$ is minimized when ${\mathbb{P}}_{gd}(x,y)={\mathbb{P}}_{dd}(x,y)$

\subsubsection{Dual Discriminator Loss Function of cGAN}

We propose a three player game with two discriminators $D_{xy}$ and $D_{x}$ and one generator. The disriminator $D_{xy}$ sees the inputs $x$ and $y$, whereas $D_x$ sees only $x$. The new objective function becomes

\begin{eqnarray}
\min_G\max_{D_{xy},D_{x}} & \mathbb{E}_{y\sim {\mathbb{P}}_d(y)}\mathbb{E}_{x\sim {\mathbb{P}}_d(x\mid y)}\log\left[D_{xy}\left(x,y\right)\right]\\ \nonumber
&+\mathbb{E}_{y\sim {\mathbb{P}}_d(y)}\mathbb{E}_{\widehat{x}\sim {\mathbb{P}}_g(x\mid y)}\log\left[1-D_{xy}\left(\widehat{x},y\right)\right] \\ \nonumber
&+\mathbb{E}_{x\sim {\mathbb{P}}_d(x)}\log\left[D_{x}\left(x\right)\right] \\ \nonumber
&+\mathbb{E}_{\widehat{x}\sim {\mathbb{P}}_g(x\mid y)}\log\left[1-D_{x}\left(\widehat{x}\right)\right] \nonumber
\end{eqnarray}

Expanding as before, we get

\begin{eqnarray}
&\displaystyle \int_y{{\mathbb{P}}_d(y)\int_x{{\mathbb{P}}_d(x\mid y)}\log[D_{xy}(x,y)]}\\ \nonumber
+&\displaystyle \int_y{{\mathbb{P}}_d(y)}\int_x{{\mathbb{P}}_g(x\mid y)\log[1-D_{xy}(x,y)]}\\ \nonumber
+&\displaystyle \int_x{{\mathbb{P}}_d(x)}\log[D_x(x)]+\displaystyle \int_x{{\mathbb{P}}_g(x\mid y)\log[1-D_x(x)]} \nonumber
\end{eqnarray}

By differentiating and equating to zero, we get the optimum values of $D^*_{xy}(x,y)$ and $D^*_{x}(x)$ as Eq. \ref{AD_2} and as follows respectively

\begin{equation}
D^*_{x}\left(x\right)=\frac{{\mathbb{P}}_d(x)}{{\mathbb{P}}_d(x)+{\mathbb{P}}_{g}(x/y)}
\end{equation}

Substituting this optimal values $D^*_{xy}(x,y)$ and $D^*_{x}(x)$ into the generator optimization function, we get

\begin{eqnarray}
\min_G &\displaystyle \int_y{{\mathbb{P}}_d(y)\int_x{{\mathbb{P}}_{dd}(x\mid y)}\log\left[\frac{{\mathbb{P}}_{dd}(x,y)}{{\mathbb{P}}_d(x,y)+{\mathbb{P}}_{gd}(x,y)}\right]}\\ \nonumber
+&\displaystyle \int_y{{\mathbb{P}}_d(y)}\int_x{{\mathbb{P}}_g(x\mid y)\log\left[\frac{{\mathbb{P}}_{gd}(x,y)}{{\mathbb{P}}_{dd}(x,y)+{\mathbb{P}}_{gd}(x,y)}\right]}\\ \nonumber
+&\displaystyle \int_x{{\mathbb{P}}_d(x)}\log\left[\frac{{\mathbb{P}}_d(x)}{{\mathbb{P}}_d(x)+{\mathbb{P}}_{g}(x/y)}\right]\\ \nonumber
+&\displaystyle \int_x{{\mathbb{P}}_g(x\mid y)\log\left[\frac{{\mathbb{P}}_{g}(x/y)}{{\mathbb{P}}_g(x)+{\mathbb{P}}_{g}(x/y)}\right]} \nonumber
\end{eqnarray}

On simplification (as done for conventional conditional GAN), the above equation reduces to

\begin{eqnarray}
&-4\log(4)+KL\left[{\mathbb{P}}_{dd}(x,y)||\frac{{\mathbb{P}}_{dd}(x,y)+{\mathbb{P}}_{gd}(x,y)}{2}\right]\\ \nonumber
&+KL\left[{\mathbb{P}}_{gd}d(x,y)||\frac{{\mathbb{P}}_{dd}(x,y)+{\mathbb{P}}_{gd}(x,y)}{2}\right]\\ \nonumber
&+KL\left[{\mathbb{P}}_d(x)||\frac{{\mathbb{P}}_d(x)+{\mathbb{P}}_{g}(x\mid y)}{2}\right]+KL\left[{\mathbb{P}}_{g}(x\mid y)||\frac{{\mathbb{P}}_{d}(x)+{\mathbb{P}}_{g}(x\mid y)}{2}\right] \nonumber
\end{eqnarray}

The last four terms turns out to be sum of $JSD({\mathbb{P}}_{dd}(x,y)||{\mathbb{P}}_{gd}(x,y))$ and $JSD({\mathbb{P}}_d(x)||{\mathbb{P}}_{g}(x\mid{y}))$. Hence generator achieves its minimum when $p_{g}(x\mid{y})=p_d(x)$ and $p_{gd}(x,y)=p_{dd}(x,y)$.

cGAN achieves optimum setting when $p_{gd}(x,y)=p_{g}(x\mid{y})p_d(y)=p_{dd}(x,y)$, whereas dual discriminator cGAN enforces a constraint on the $p_{g}(x\mid{y})$, which enables it to memorize the objects in the training set in addition to learning input-output image relation.

\section{Experimental Validation}
In this section, we provide the details of the dataset and the experiments conducted to validate the proposed algorithm

\subsection{Dataset}

Although few event-based datasets are available for other vision-based tasks such as visual odometry \cite{EMuegglerandHRebecq}, object recognition \cite{HLiandHLiuandXJi}, there is no event-based dataset available for anomaly detection. To address this constraint and in order to evaluate the proposed algorithm, we introduce a new event dataset for anomaly detection. We present two variations of event dataset from two distinctive sets of environments, indoor lab environment, and outdoor corridor environment to set a realistic baseline for algorithm evaluation. This dataset is comprised of short event clips of pedestrian movements parallel to the camera plane, captured from static DAVIS camera with a resolution of $346\times{260}$ . The normal and anomalous scenes are staged with typical training videos consisting of people walking, talking, sitting on a couch etc. The testing videos consist of the following variety of anomalous activities: people running, fighting, bending, and stealing bag. We summarize the statistics of the collected dataset in Table \ref{Table}.

\begin{table}
\centering
\begin{tabular}{|c|c||c|c|} 
\hline
\multicolumn{2}{|c||}{\textbf{Indoor}} & \multicolumn{2}{c|}{\textbf{Outdoor}}  \\ 
\hline
\hline
 \textbf{Normal} & \textbf{Instances}                        & \textbf{Normal}  &  \textbf{Instances}                           \\
 \hline
Walking &         25                & Walking  &               23              \\
 \hline
Sitting on sofa &       11                  & Sitting on chair  &    3                         \\
 \hline
 Talking &            5             & Talking  &                             \\
 \hline
 Handshaking &  & Handshaking & \\
 \hline
 \hline
 \textbf{Abnormal} & \textbf{Instances}                        & \textbf{Abnormal}  & \textbf{ Instances}                           \\
\hline

 Running &          11               &  Running &         13                    \\
 \hline
 Bending &          13              & Bending  &      13                       \\
 \hline
 Fighting &          3            & Fighting  &                             \\
 \hline
 Stealing bag &      3                  & Stealing some object  &                             \\
\hline
\end{tabular}
\caption{Details of the Normal and Anomaly videos captured in indoor and outdoor environment}
\label{Table}
\end{table}

\subsection{Evaluation Procedure}

In this section, we evaluate different components of the proposed method on the event anomaly dataset proposed in this paper. We have conducted three different sets of experiments adapted to validate DL memory surface network, dual discriminator cGAN.

\subsubsection{Experiment A: DL Memory Surface}

\begin{figure}
\centering
\includegraphics[width=\linewidth,height=1 in]{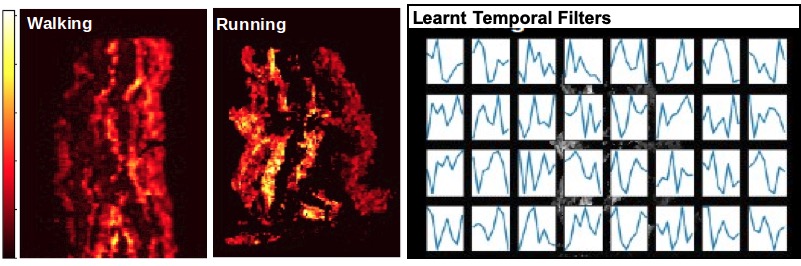}
\caption{Leftmost: DL memory surfaces generated at bottleneck layer of DL memory surface network on walking (first) and running (second) sequences, Rightmost: Visualization of 32 encoder bases learnt by DL memory surface network}
\label{fig:DL_mem}
\end{figure}

In this segment, we shed light on the qualitative assessment of our proposed DL memory surface generation network on real event data. We applied the DL network on a training set of discretized event volumes of two different temporal classes, walking and running. Towards qualitative analysis, we provide the visualization (Fig. \ref{fig:DL_mem}) of DL memory surfaces (of walking and running) and the temporal filters learnt by the network. The filters learned by the algorithm represent the different motion models of the events presented to the network.

 \subsubsection{Experiment B: Dual Discriminator cGAN}

This section is dedicated to analysis of the dual discriminator cGAN in terms of anomaly detection subsequent to event prediction. Towards this, a training set of normal activity clips such as pedestrians walking, talking, sitting, etc are provided to the network for learning the model. 

\begin{figure}

  \centering
  \includegraphics[width=\linewidth,height=2 in]{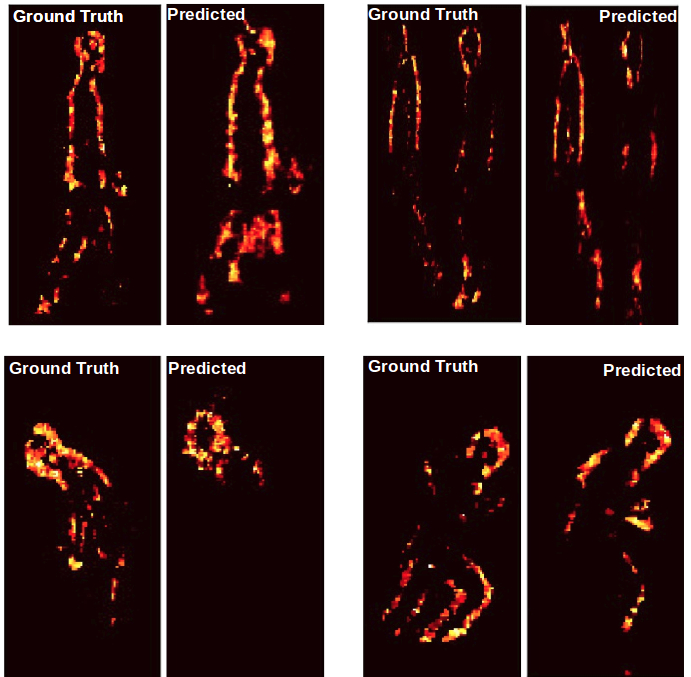}  
\caption{Top row: Person walking, Bottom row: Anomalies (Left: Person bending and Right: running). The system predicts the normal activity (top row), whereas the prediction accuracy is low in the case of anomalies}
\label{fig:recons}
\end{figure}

 As part of qualitative validation, we provide the prediction output of the network for normal activity and abnormal activity in Fig. \ref{fig:recons}. It is evident that the prediction capability of the network is well pronounced for normal activities than that of anomalous activities.

\begin{figure}[h]
\centering
\includegraphics[width=\linewidth,height=2 in]{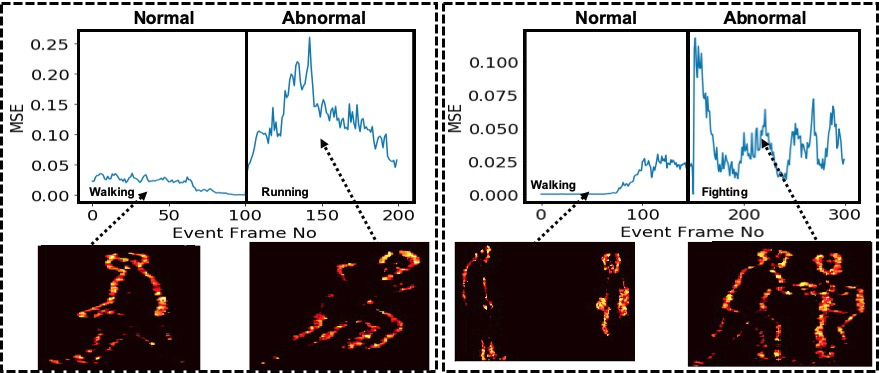}  
\caption{Plot of MSE between original and predicted frames. Left: Sequence with walking as normal activity and running as abnormal activity. Right: Sequence with walking as normal activity and fighting as abnormal activity}

\label{fig:MSE}

\end{figure}

So as to quantitatively evaluate the performance of the anomaly detection efficiency of dual discriminator cGAN, testing cases with intermittent events of abnormal activities such as running and fighting, etc were presented to the network. The proposed algorithm detects the presence of anomalous events at the event frame level by evaluating Mean Square Error (MSE) between the predicted events and the ground truth events. It should be noted that the co-occurrence of the spatial location of anomaly is not considered for evaluation. Fig. \ref{fig:MSE} shows the plot of MSE vs event frame number. It can be seen that MSE is higher for events such as running and fighting resulting from the fact that these kind of motion never appeared in the training set. Running has been distinguished as anomaly with higher probability than that of fighting sequence.

\section{Conclusion}
In this paper, we presented the first baseline for event-based anomaly detection by potentially taking advantage of the most well-known deep learning models such as generative model and encoder-decoder model. The proposed solution involves cGAN with dual discriminator loss function, which permits capture of the spatial anomaly as well in event domain that might have gotten away from the cGAN with single discriminator. We have also proposed a first in the line deep learning solution to effectively encode the event data as sparse DL memory surface, where-in the motion information introduced by the event cameras at the sensor level is learnt adaptively from the data as opposed to state-of-the-art hard-wired event data representations. We have also provided an event-based anomaly dataset on which the proposed algorithm has been validated from different perspectives. It has also been displayed that the proposed method utilizes the motion information obtained from event sensor effectively resulting in huge saving in terms of computation and performance as compared to state-of-the-art conventional frame-based anomaly detection networks.


%

\ifCLASSOPTIONcaptionsoff
  \newpage
\fi



{\small
\bibliographystyle{ieee_fullname}
\bibliography{egbib}
}
%

%




\end{document}